\documentclass{article}
\PassOptionsToPackage{numbers, compress}{natbib}
\usepackage[final]{neurips_2024}
\usepackage[utf8]{inputenc} 
\usepackage[T1]{fontenc}    
\usepackage{hyperref}       
\usepackage{url}            
\usepackage{booktabs}       
\usepackage{amsfonts}       
\usepackage{nicefrac}       
\usepackage{microtype}      
\usepackage{xcolor}         
\usepackage{multirow}
\usepackage{subcaption}
\usepackage{array}
\usepackage{amsmath}
\usepackage{makecell}
\usepackage[shortlabels]{enumitem}
\usepackage{graphicx} 
\usepackage{booktabs} 
\usepackage{colortbl}
\usepackage{wrapfig}
\usepackage{picinpar}
\usepackage{cutwin}
\title{Valley2: Exploring Multimodal Models with Scalable Vision-Language Design}

\author{%
  Ziheng Wu\textsuperscript{*}, Zhenghao Chen\textsuperscript{*}, Ruipu Luo, Can Zhang,\\
  \textbf{Yuan Gao, Zhentao He, Xian Wang, Haoran Lin, Minghui Qiu} \\
  ByteDance \\
  \texttt{\href{mailto:wuheng.2024@bytedance.com}{wuheng.2024@bytedance.com}}
}

\begin{document}
\maketitle

\begin{abstract}
Recently, vision-language models have made remarkable progress, demonstrating outstanding capabilities in various tasks such as image captioning and video understanding. We introduce \textbf{Valley2}, a novel multimodal large language model designed to enhance performance across all domains and extend the boundaries of practical applications in e-commerce and short video scenarios. 
(i) We curate high-quality instruction datasets and benchmarks in the e-commerce and short video domains, featuring multimodal inputs, extended image-video sequences requiring interaction, and rich domain knowledge, enabling Valley2 to achieve state-of-the-art performance. (ii) We introduce innovations such as a large visual vocabulary, convolutional adapters, and the Eagle Module, improving flexibility in handling diverse real-world inputs (e.g., ultra-long video-image combinations, extreme aspect ratios) while enhancing training and inference efficiency. (iii) Chain-of-Thought (CoT) Post-Training further optimizes Valley2’s reasoning capabilities, boosting overall performance and revealing directions for future improvements. Notably, \textbf{Valley2} achieves state-of-the-art (SOTA) performance on e-commerce benchmarks, surpassing open-source models of similar size by a large margin (79.66 vs.\ 72.76). Additionally, Valley2 ranks second on the OpenCompass leaderboard among models with fewer than 10B parameters, with an impressive average score of 67.4. The code and model weights are open-sourced at \href{https://github.com/bytedance/Valley}{https://github.com/bytedance/Valley}.
\end{abstract}

\section{Introduction}
\label{intro}

In recent years, large language models (LLMs)~\cite{radford2018improving,brown2020language,ouyang2022training,achiam2023gpt,yang2024qwen2} have made remarkable strides, exhibiting human-level performance in numerous natural language processing (NLP) tasks and contributing significantly to the development of artificial general intelligence (AGI) systems. Meanwhile, multimodal large language models (MLLMs)~\cite{liu2024llava, chen2024far,yao2024minicpm, wang2024qwen2vl, liu2024points1, lu2024bluelm, zhang2023video, liu2024kangaroo}, which integrate LLMs with multimodal processing capabilities, have enabled advanced vision/audio-language interactions and dialogues, tackling complex tasks such as mathematical problem solving, long document understanding, and speech intent recognition.

Despite these achievements, open-source models generally fall behind proprietary systems like GPT-4~\cite{achiam2023gpt} and Claude-3.5-Sonnet in terms of overall capabilities. To narrow this gap, the open-source community has vigorously pursued a variety of innovative solutions~\cite{guo2025llavauhd, wang2024qwen2vl,chen2024far,li2024llavaov,yao2024minicpm,liu2024points1,chen2024internvl2}, achieving competitive or even superior results. This line of research emphasizes the potential of multimodal perception, wherein LLMs are extended with inputs such as images, videos, and audio via advanced cross-modal pre-training and instruction-tuning techniques.

In this paper, we introduce \textbf{Valley2}, a novel multimodal large language model that supports text, image, and video inputs. Currently, it ranks \textbf{second} on the OpenCompass leaderboard~\cite{2023opencompass} among models with fewer than 10B parameters, achieving an average score of 67.4. It also attains state-of-the-art (SOTA) performance on e-commerce benchmarks, surpassing open-source models of similar size by a considerable margin. Compared to our previous work, Valley~\cite{luo2023valley}, the latest \textbf{Valley2} explores enhancements in tiling strategies, model architecture, data construction, and training methodologies. 

To strengthen the representation ability for complex visual inputs without introducing an excessive number of visual tokens, we adopt a tiling strategy similar to previous works~\cite{liu2024llavanext,li2024llavaov, chen2024far} for single images, yet avoid its use for multi-image or video inputs. Moreover, we remove padding tokens from the vision encoder to increase the proportion of informative tokens. Inspired by Ovis~\cite{lu2024ovis}, we introduce a learnable visual vocabulary to capture richer semantic information from visual data. PixelShuffle~\cite{shi2016real}, designed as a channel-to-space transformation, is used to reduce token count while retaining information. However, under the design of a visual vocabulary with a larger hidden dimension, PixelShuffle considerably expands the number of parameters. To address this, Valley2 adopts a lightweight ConvAdapter consisting of two MLP layers and one convolutional layer; the convolutional layer performs 2$\times$ down-sampling along both spatial dimensions, ensuring our large MLP design is not constrained by an excessive parameter count. Furthermore, to handle extreme-resolution images and deliver superior performance, Valley2 introduces the Eagle Module~\cite{shi2024eagle} which adds an additional vision encoder in parallel and concatenates the vision tokens from both encoders along the sequence dimension. Leveraging heterogeneous vision encoders and a larger number of visual tokens, our Valley2 model achieves competitive performance on both OpenCompass and in-house e-commerce benchmarks.

\section{Model Architecture}
\subsection{Overview}
\begin{figure}[htbp]
    \centering
    \hspace*{-0.09\textwidth} %
    \makebox[\textwidth]{\includegraphics[width=1.2\textwidth]{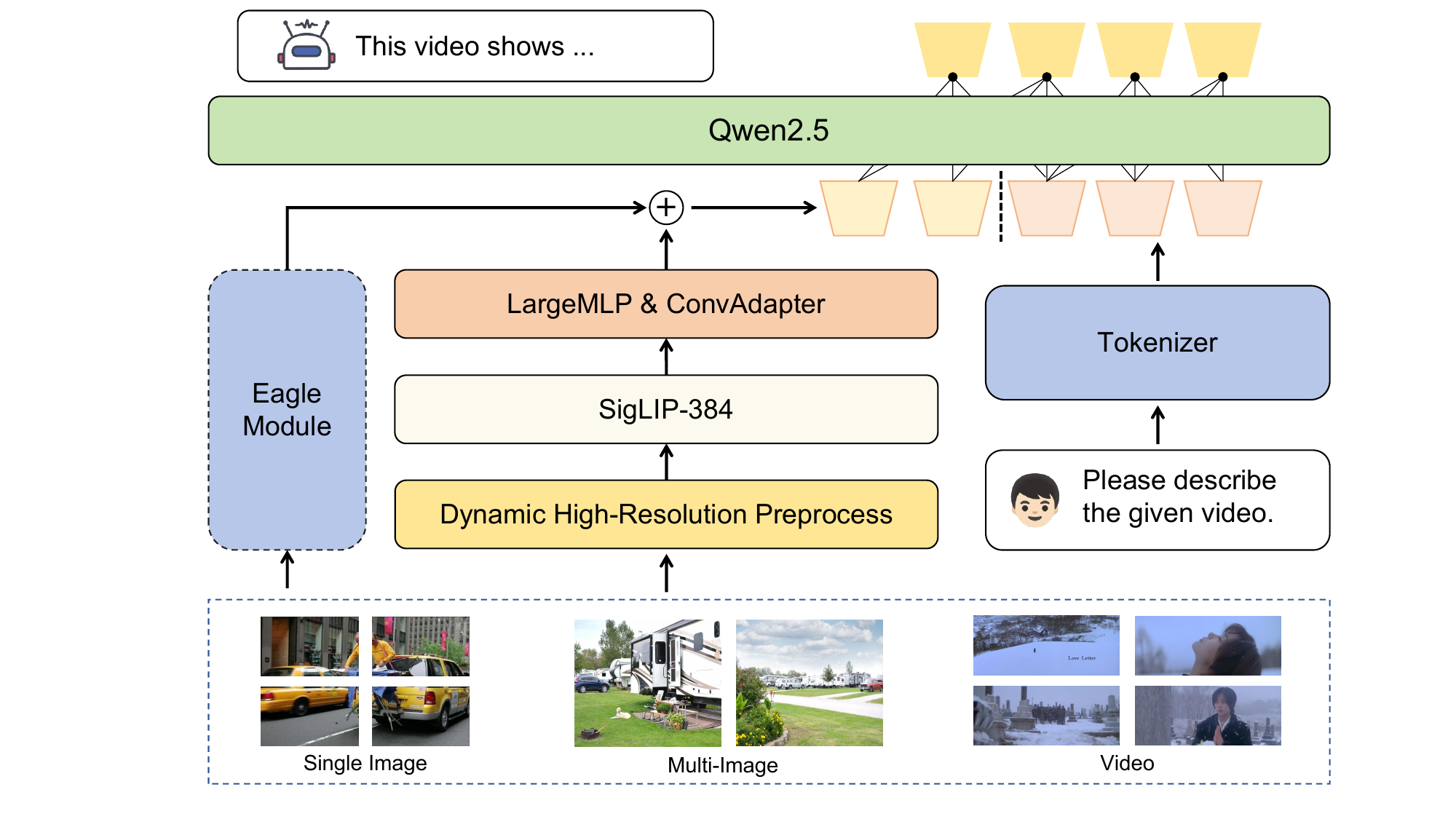}}
    \caption{Overview of Valley}
    \label{fig:overview}
\end{figure}
\vspace{-0.3cm}

The structure of Valley2 is shown in Figure~\ref{fig:overview}. Valley2 adopts Qwen2.5\cite{yang2024qwen2} as its LLM backbone and SigLIP-384\cite{zhai2023sigmoid} as the vision encoder, alongside a projector based on a combination of MLP layers and convolution for efficient feature transformation.

Valley2 employs a high-resolution tiling strategy inspired by InternVL2\cite{chen2024far}, using a best-ratio matching mechanism to maintain aspect ratio consistency. We restrict this strategy tiling for single images to a maximum of 9 slices, whereas video data and multi-image data bypass tiling entirely to control the total token count. Unlike previous designs involving AnyRes ~\cite{li2024llavaov} and PixelShuffle, Valley2 uses a simple convolution operation to downsample visual tokens by a factor of 2 while preserving the original dimension.

\subsection{Projector: Large MLP and ConvAdapter}

The initial design of the projector was inspired by Ovis, adopting a two-layer MLP with enlarged hidden dimensions to enhance feature representation. Ablation studies confirmed notable performance gains, especially when paired with fixed-vision encoders such as SigLIP. 

However, the use of PixelShuffle as a space-to-dimension operation posed significant constraints. While PixelShuffle effectively leveraged the spatial locality of visual features to reduce tokens without sacrificing information, combining it with larger hidden dimensions led to a substantial increase in projector parameters, causing training instability and limiting scalability in multimodal tasks. To address these challenges, Valley2 introduces a lightweight ConvAdapter, which reduces the token count of vision encoder outputs without expanding their dimensions. This approach significantly lowers the parameter count in the larger MLP design. ConvAdapter preserves robust performance while enhancing both training stability and inference efficiency, marking a major step forward in model optimization.

\subsection{Eagle Module: Addressing Distortions and Extending Token Representation}
\label{subsec:eagle-module}

Valley2’s aggressive token compression strategy revealed performance limitations in handling extreme input scenarios, notably in OCR tasks and large-document understanding. The tiling approach often struggled to reconcile the conflicting demands of fine-grained document processing and large-scale visual inputs in mixed text-video or multi-image contexts. On the one hand, these shortcomings stemmed from tiling methods inherited from InternVL2, which introduced image distortions and struggled with extreme aspect ratios; on the other hand, the insufficient resolution and token capacity designed for one-vision tasks further exacerbated these issues in demanding scenarios.

To overcome these constraints, Valley2 incorporates the Eagle Module~\cite{shi2024eagle}. This module uses a parallel vision encoder to reduce distortions while ensuring compatibility with extreme inputs, native aspect ratios, and efficient training and inference. During training, the token output of the additional vision encoder is constrained to match the number of tokens produced by the tiling strategy, avoiding excessive overhead and maintaining balanced computational costs. During inference, these constraints are relaxed, allowing the Eagle branch to adapt token counts flexibly based on specific input requirements. This dynamic design increases versatility in both high-resolution and low-resolution applications.

\subsection{Summary}
Valley2 leverages Qwen2.5, one of the most advanced language backbones, to achieve superior performance in multimodal tasks. Its design incorporates multiple optimizations for images and videos:

\begin{enumerate}[label={\bf {{$\bullet$}}},,leftmargin=*,topsep=0.5ex,itemsep=-0.5ex,partopsep=0.75ex,parsep=0.75ex,partopsep=0pt,wide,labelindent=0pt]
    \item \textbf{High-Resolution Tiling.} Employs a carefully designed high-resolution tiling strategy, effectively regulating token counts across various modalities while delivering strong results without compromising efficiency.

    \item \textbf{Large Projector with ConvAdapter.} Combines a ConvAdapter with a large hidden-layer MLP, replacing the previous PixelShuffle-based approach to produce a high-performance, efficient, and stable training projector.

    \item \textbf{Eagle Module.} Alleviates tiling-induced distortions and precision trade-offs by dynamically adapting to extreme input cases, thereby enhancing scalability and robustness.
\end{enumerate}

With these optimizations, Valley2 achieves an average training sequence length of under 1{,}000 tokens, as well as a scalable maximum token range of about 1{,}000$\sim$2{,}000 for single-image inference matching the performance of models that require 4{,}000$\sim$8{,}000 tokens per image. This balance ensures that Valley2 remains both efficient and highly competitive for real-world applications.

\section{Data}

In this section, we introduce the three components of the Valley2 data: (1) OneVision-style data for each training phase of multimodal large models, (2) data and evaluation specifically targeted at e-commerce and short-video domains, and (3) construction of Chain-of-Thought (CoT) data tailored for complex problem-solving. By combining these three types of data, we created a model that not only delivers exceptional performance on general multimodal tasks in OpenCompass but also excels in e-commerce content understanding.

\subsection{OneVision Data}
\subsubsection{Text-Vision Aligning Data}
We collect text-vision aligning data generated by Emu2~\cite{Emu2} and integrate them with other high-quality open-source caption datasets to form a 7.5M large-scale dataset. This caption dataset not only provides effective support for the initial training of large-scale multimodal models but also establishes a foundational understanding of image-text relationships.

\subsubsection{High-Quality Knowledge Learning Data}
Following the training strategy of LLaVA-Next~\cite{liu2024llavanext}, we meticulously construct a High-Quality Knowledge Learning Dataset. This dataset significantly enhances the model’s cross-modal understanding and reasoning capabilities by incorporating multimodal, multitask data. In specific task domains, we systematically integrate and optimize data sources, including image captioning (CC3M~\cite{liu2024llava}), image localization (Flickr30k~\cite{plummer2015flickr30k}, GRIT~\cite{kakaobrain2022coyogrit-700m}, BLIP3-GROUNDING-50M~\cite{blip3-xgenmm}), OCR (BLIP3-OCR-200M~\cite{blip3-xgenmm}), interleaved image-text (MMC4~\cite{zhu2023mmc4}), and video captioning (OpenVid-1M, vript). 

\subsubsection{Visual Instruction Data}
Table~\ref{tab:visual_instruction_data} summarizes the various types of visual instruction data integrated into our training pipeline\footnote{Datasets in the table have been sampled.}, including details on the relevant datasets and the objectives they serve. This structured overview clarifies how each category of tasks—ranging from general to OCR-specific—contributes to enhancing the model’s multimodal capabilities.
\vspace{-0.3cm}
\begin{table}[htbp]\small
\centering
\renewcommand{\arraystretch}{1.2}
\setlength{\tabcolsep}{4pt} 
\caption{Overview of Visual Instruction Data}
\label{tab:visual_instruction_data}
\resizebox{\textwidth}{!}{ 
\begin{tabular}{l p{10cm} c} 
\toprule
\textbf{Type} & \textbf{Datasets} & \textbf{Size} \\ 
\midrule

\textbf{General Task} 
& 
Vision-Flan~\cite{xu2024vision},
LLaVA-158K~\cite{liu2023visual},
VQA-RAD~\cite{lau2018vqa},
ST-VQA~\cite{biten2019scene},
Hateful Memes~\cite{kiela2020hateful},
RefCOCO~\cite{yu2016modeling},
TallyQA~\cite{acharya2019tallyqa},
ShareGPT4V~\cite{chen2023sharegpt4v},
CLEVR~\cite{johnson2017clevr},
IconQA~\cite{lu2021iconqa},
LLaVAR~\cite{liu2024llavar},
COCO Caption~\cite{lin2014microsoft},
VQAv2~\cite{goyal2017making},
ScienceQA~\cite{lu2022scienceqa},
ALLaVA Inst~\cite{chen2024allava},
Image Textualization~\cite{pi2024image},
VizWiz~\cite{gurari2018vizwiz}, AOKVQA~\cite{marino2021okvqa}, LLaVA-Wild~\cite{liu2023llava}, 
Visual7W~\cite{zhu2016visual7w}, VSR~\cite{liu2022visual}, Coco Caption~\cite{lin2014microsoft}, OKVQA~\cite{marino2019okvqa},
ShareGPT4o, WebSight, InterGPS, Inhouse Ecom-Data
&
6.35M 
\\

\textbf{Reasoning Task}
&
Geo170K Align~\cite{glava2024},
CLEVR-Math~\cite{lindstrom2022clevr},
Geometry3K~\cite{lu2023intergps},
Super-CLEVR~\cite{li2023super},
GQA~\cite{hudson2019gqa},
Geo170K~\cite{glava2024}, RAVEN~\cite{zhang2019raven},
Visual Genome~\cite{krishna2017visual},
MathQA~\cite{roy2019mathqa},
MAVIS Data Engine~\cite{zhang2024mavis},
Geometry3K~\cite{lu2023intergps}, 
TabMWP~\cite{lu2022dynamic}, 
MathV360K~\cite{bin2024mathllava}, 
LRV Normal~\cite{liu2024aligning}, 
UniGeo~\cite{zhang2023unimath}, 
Infinity-MM~\cite{gu2024infinitymm}, Inhouse Ecom-Data
&
1.8M
\\ 

\textbf{Pure-text Task}
&
Magpie Pro~\cite{magpie2024alignment}, Infinity-Instruct~\cite{zhang2024infinity}
&
2.0M
\\ 

\textbf{OCR Task}
&
Rendered Text, 
TextCaps~\cite{TextCaps2020},
IAM~\cite{IAM2002},
HME100K~\cite{Yuan2022},
TextOCR~\cite{Singh2021},
IIT5K~\cite{Mishra2012},
OCRVQA~\cite{Singh2019},
SynthDog-EN~\cite{Kim2022},
CROHME~\cite{Mouchere2019}
, Inhouse Ecom-Data
&
0.7M
\\ 

\textbf{OneVision Task}
&
Inhouse Ecom-Data, including multi-images and videos
&
0.65M
\\

\bottomrule
\end{tabular}}
\end{table}

\subsection{E-commerce Data and Benchmark}

Existing open-source benchmarks can be categorized based on their focus. Knowledge-based benchmarks (e.g., MMMU~\cite{yue2023mmmu}, MathVista~\cite{lu2024mathvista}, and AI2D~\cite{kembhavi2016diagramai2d}) emphasize general or domain-specific knowledge evaluation. Vision and OCR benchmarks (e.g., OCR-Bench~\cite{ocrbench}, ChartQA~\cite{masry-etal-2022-chartqa}, RealWorldQA) primarily target visual perception and textual recognition. Single-image multimodal benchmarks, represented by MMVet~\cite{yu2024mmvet} and MMBench~\cite{MMBench}, assess the combination of multiple capabilities within single-image tasks. Finally, video benchmarks like VideoMME~\cite{fu2024videomme} focus on reasoning and understanding over video inputs.

Short-video e-commerce data, exemplified by E-com, present unique characteristics beyond the scope of these benchmarks. From a knowledge perspective, E-com data involve domain-specific knowledge that pairs unique visual requirements with specialized common-sense reasoning. Regarding data format, E-com datasets incorporate text, images, video, and audio within a single instance, demanding comprehensive multimodal understanding and decision-making—aligning with OneVision tasks. Moreover, reasoning in E-com benchmarks often involves multiple judgments and exceptions, necessitating systematic reasoning capabilities.

To address these challenges, we construct and annotate specialized datasets for e-commerce evaluation:

\textbf{Ecom-Caption Benchmark.} Ecom-Caption Benchmark is a multimodal alignment dataset derived from open-source models for e-commerce-related images and videos. It primarily evaluates text-vision alignment during the initial training stage. With 5{,}592 samples, the benchmark’s evaluation metrics include BLEU-2, CIDEr, METEOR, and ROUGE\_L. BLEU-2 measures the precision of bigram overlaps between generated captions and ground-truth references.

\begin{figwindow}[0,r,
{\includegraphics[width=7cm]{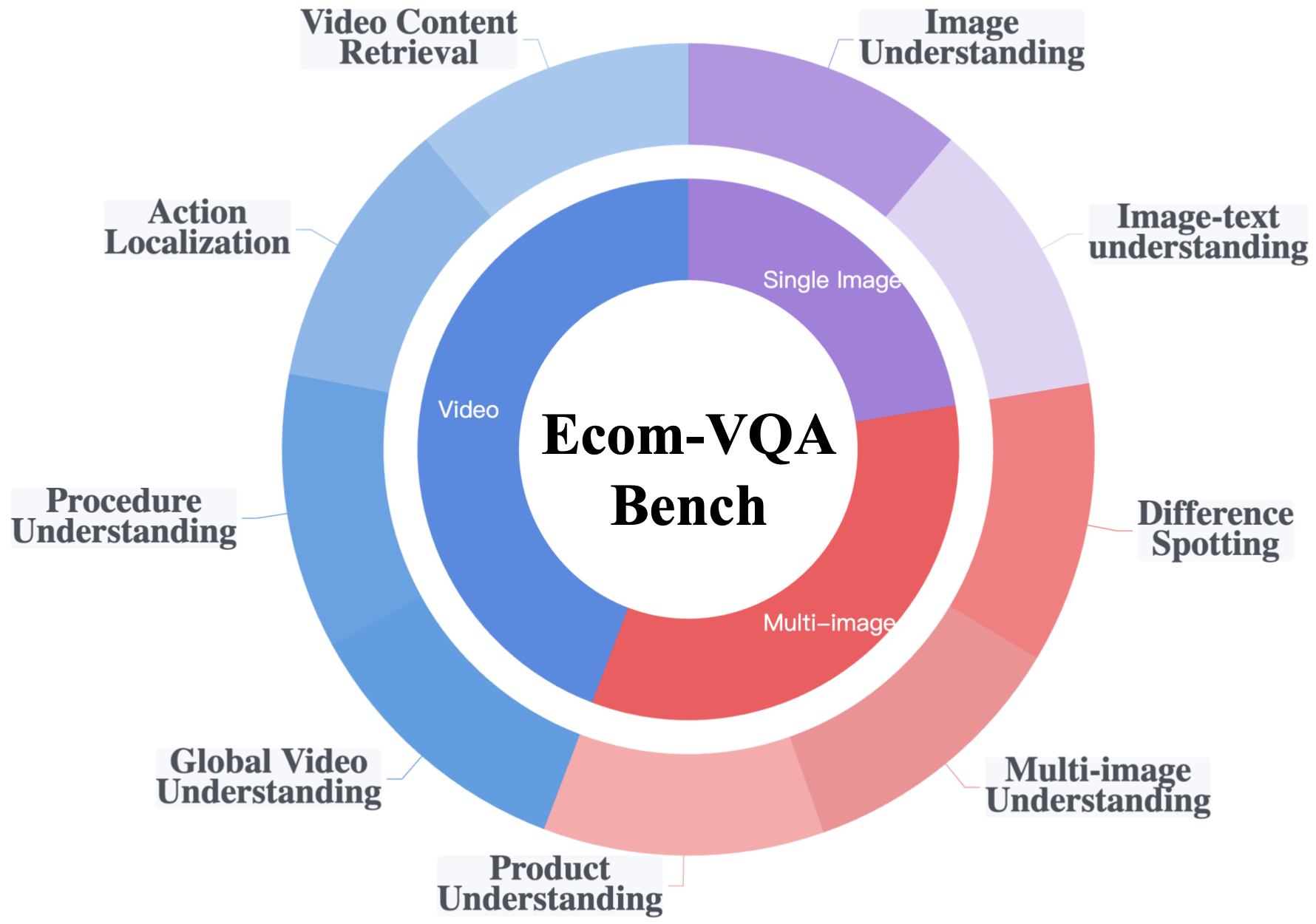}},
{The data distribution of Ecom-VQA Bench}]

\textbf{Ecom-VQA Benchmark.} To assess the model’s understanding of e-commerce domain knowledge and multimodal reasoning ability, we collect e-commerce images and videos to build the Ecom-VQA Benchmark. It comprises 536 QA pairs across three modalities: single image, multiple images, and video. Table~\ref{tab:Ecom-VQA} shows the distribution for each modality. The benchmark uses a multiple-choice format with four options, and the model must select the correct answer (A, B, C, or D). We evaluate multi-choice accuracy via a cyclic testing strategy: each question is repeated four times with the correct answer rotated through positions A, B, C, and D. One question is passed only if all four predictions are accurate.
\end{figwindow}

To robustly address these benchmarks and validate our model’s capabilities, we build a large-scale training corpus consisting of 2M multimodal alignment data, 5M knowledge-injected instances (covering images, videos, OCR tasks, and localization challenges), and 1M high-quality instruction data. These datasets are evenly distributed across training phases, with the proposed benchmarks used to track and assess the development of the corresponding capabilities.

\subsection{Enhancing Multimodal Model Capabilities through Systematic Reasoning}
\begin{figure}[h]
    \centering
    \includegraphics[width=1\textwidth]{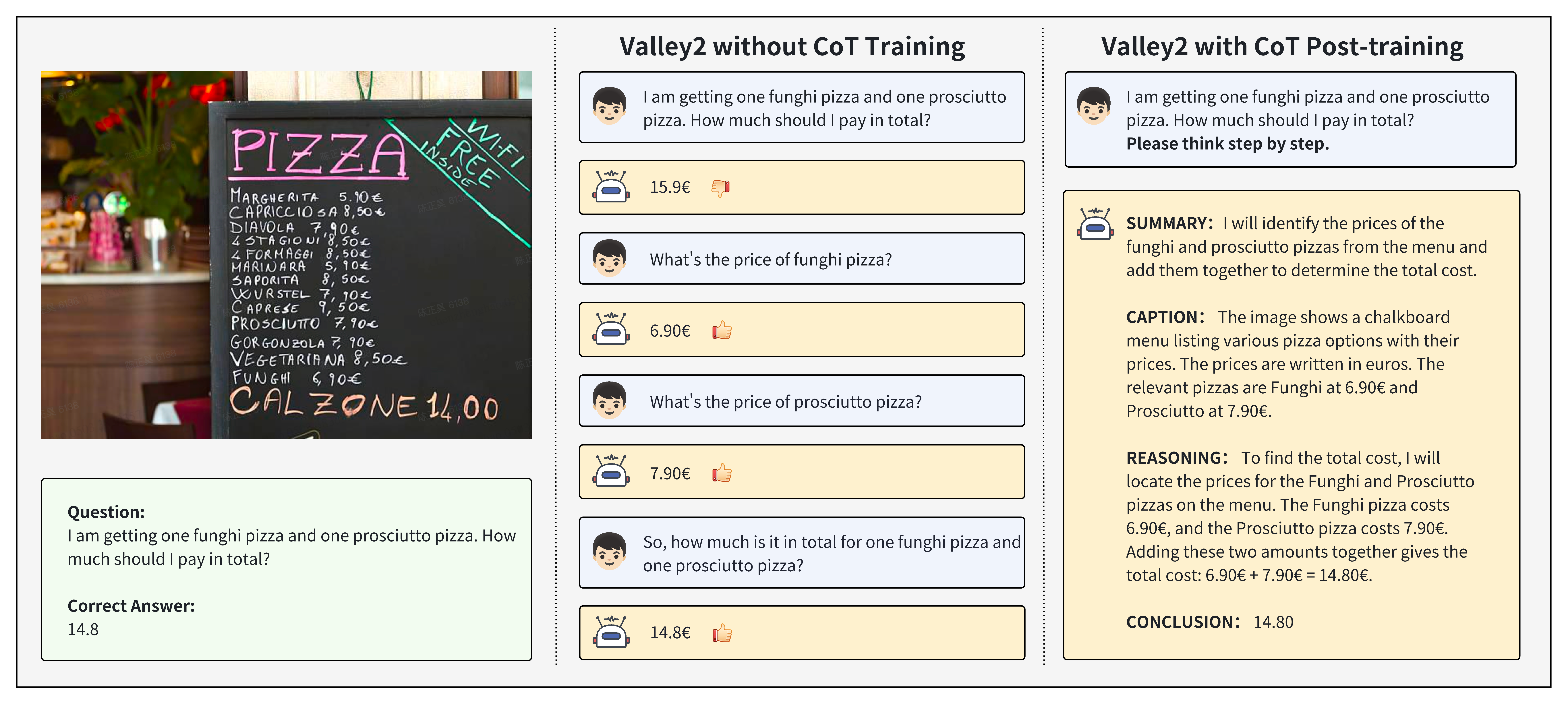}
    \caption{Comparison of Problem-Solving Approaches With and Without Chain-of-Thought Training}
    \label{fig:cot_challenge}
\end{figure}

Large multimodal models demonstrate three core competencies: visual perception, logical reasoning, and instruction-following. Yet, as illustrated by the challenging case in Figure~\ref{fig:cot_challenge}, even if a model possesses all three capabilities independently, an incorrect “thought” process may still lead to a wrong final answer.

\vspace{-0.3cm}

To address this issue, we integrate Chain-of-Thought (CoT) data so the model learns to systematize its analysis and outputs. By introducing high-quality instructional data, this approach bolsters the model’s ability to generate structured outputs. Referring to~\cite{reasoner}, we use two data types during the final Post-Training phase. The first type is CoT instruction data, where each query ends with the special prompt “Please think step by step.” The second type is non-CoT instruction data. By mixing these two types 1:1, the model acquires systematic reasoning abilities while also retaining the ability to respond directly. These modes can be switched seamlessly with a simple system prompt.

Specifically, we collect 100k CoT samples from the LLAVA-CoT dataset~\cite{xu2024llavacot, dong2024insight, reasoner} and 100k non-CoT samples randomly selected from the high-quality instruction data of Stage~2 ~\cite{gu2024infinity, li2024llavaov}. This balanced dataset spans various tasks—math, chart comprehension, document analysis, and other complex reasoning challenges. Each CoT sample contains four reasoning steps (summary, caption, reasoning, and conclusion), which substantially improve the model’s capacity to conduct systematic analysis.

\section{Training}
\subsection{Training Scheme}

We adopt a structured training pipeline widely used in multimodal large models, such as LLaVA~\cite{liu2024llava}, InternVL2~\cite{chen2024internvl2}, and Points~\cite{liu2024points1}. This pipeline consists of three primary stages: \textbf{Text-Vision Aligning (Stage-1)}, \textbf{High-Quality Knowledge Learning (Stage-1.5)}, and \textbf{Instruction Fine-Tuning (Stage-2)}. At each stage, we incorporate in-house e-commerce datasets, ensuring consistent improvements in the relevant domain capabilities while keeping the proportion of domain-specific data strictly controlled.

\begin{figure}[h]
    \centering
     \includegraphics[width=1\textwidth]{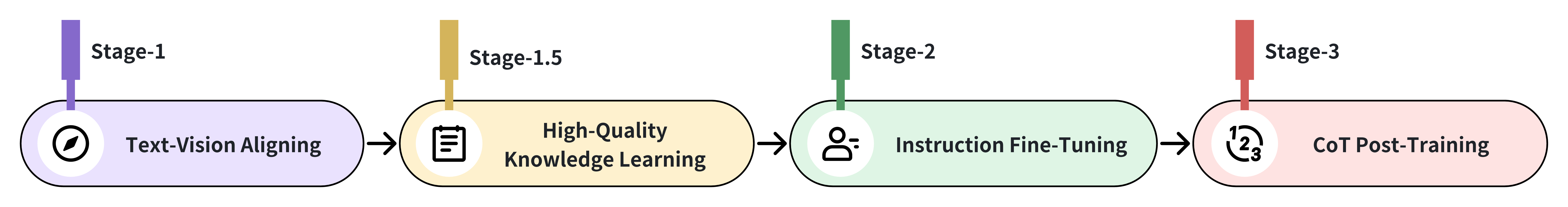}
    \caption{Training Stages}
    \label{fig:training_stages}
\end{figure}
\vspace{-0.3cm}

\begin{table}[htp]
   
    \centering
    \setlength{\tabcolsep}{12pt}
    \renewcommand{\arraystretch}{1.3}
    \resizebox{\textwidth}{!}{
    \begin{tabular}{@{}c|c|c|c|c}
    \toprule
     & \textbf{Stage-1} & \textbf{Stage-1.5} & \textbf{Stage-2} & \textbf{Stage-3 (CoT)} \\ 
    \midrule 
    \textbf{Resolution}  & 
    384 & 
    384\footnotesize{$\times$\{\{1$\times$1\}, $\cdots$, \{3$\times$3\}\}} & 
    384\footnotesize{$\times$\{\{1$\times$1\}, $\cdots$, \{3$\times$3\}\}} & 
    384\footnotesize{$\times$\{\{1$\times$1\}, $\cdots$, \{3$\times$3\}\}} \\
    \textbf{\#Tokens} & 
    Max 196\footnotesize{$\times$10} + (EAGLE) &
    Max 196\footnotesize{$\times$10} + (EAGLE) &
    Max 196\footnotesize{$\times$10} + (EAGLE) &
    Max 196\footnotesize{$\times$10} + (EAGLE) \\
    \midrule 
    \textbf{Dataset} & 7.5M & 8M & 11.5M & 0.2M \\
    \midrule
    \textbf{Trainable} & Projector & Projector+LLM & Projector+LLM & Projector+LLM \\
     \textbf{Parameter} & 163M & 7.8B & 7.8B & 7.8B \\
    \midrule 
    \textbf{Batch Size} & 96 & 96 & 192 & 192 \\
    \textbf{LR} & 1$\times10^{-4}$ & 1$\times10^{-5}$ & 1$\times10^{-5}$ & 2$\times10^{-6}$ \\
    \textbf{Epoch} & 1 & 1 & 1 & 1 \\
    \bottomrule
    \end{tabular}
    }
    \vspace{1mm}
    \caption{Detailed configuration for each training stage of Valley2.}
    \label{tab:training_strategy}
\end{table}
\vspace{-0.3cm}

As highlighted in our earlier data analysis, certain complex multimodal tasks benefit from systematic and accurate reasoning. Thus, we introduce \textbf{Chain-of-Thought (CoT) Post-Training} in the final phase \textbf{(Stage-3)} to enable the model to switch to a CoT reasoning mode. Under this mode, the model follows a fixed analytic process to produce structured outputs, leading to substantial performance gains across diverse benchmarks.

Consequently, the complete training process includes four stages. Detailed learning rates and parameter distributions are provided in Table~\ref{tab:training_strategy}. To further optimize performance and efficiency, we incorporate advanced techniques such as Packing and Annealing. Described below, these methods address critical challenges in training efficiency and robust optimization, ensuring superior performance in multiple benchmarks.

\subsection{Packing}
\begin{figure}[htbp]
    \centering
    \begin{subfigure}[t]{0.48\textwidth}
        \centering
        \includegraphics[width=6.4cm]{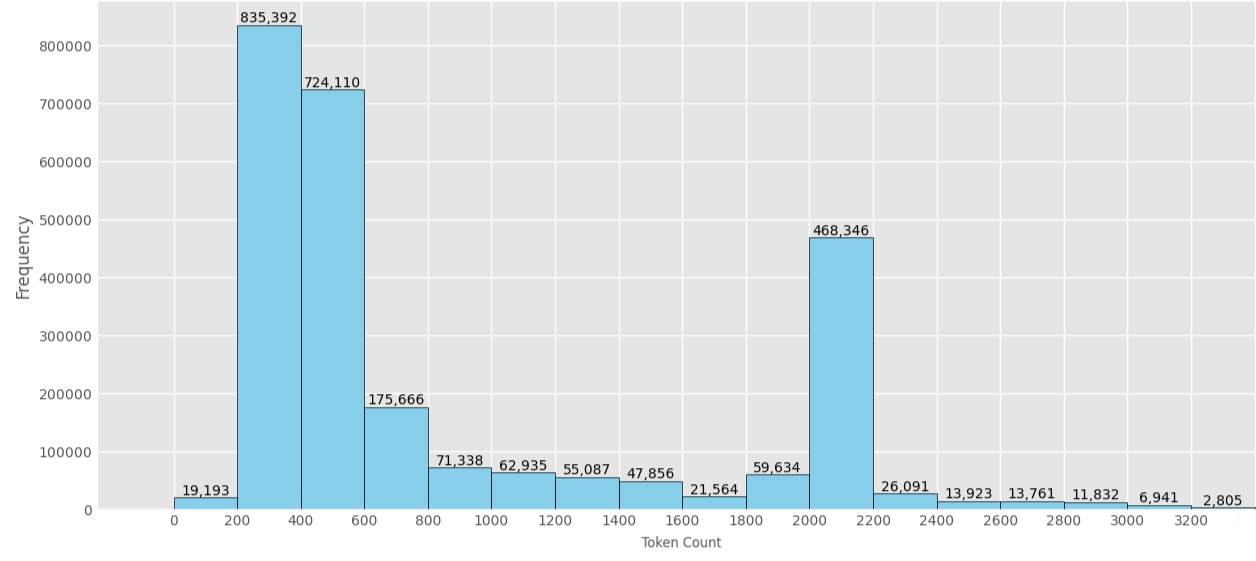}
        \caption{Before packing}
        \label{fig:before_packing}
    \end{subfigure}
    \begin{subfigure}[t]{0.48\textwidth}
        \centering
        \includegraphics[width=6.5cm]{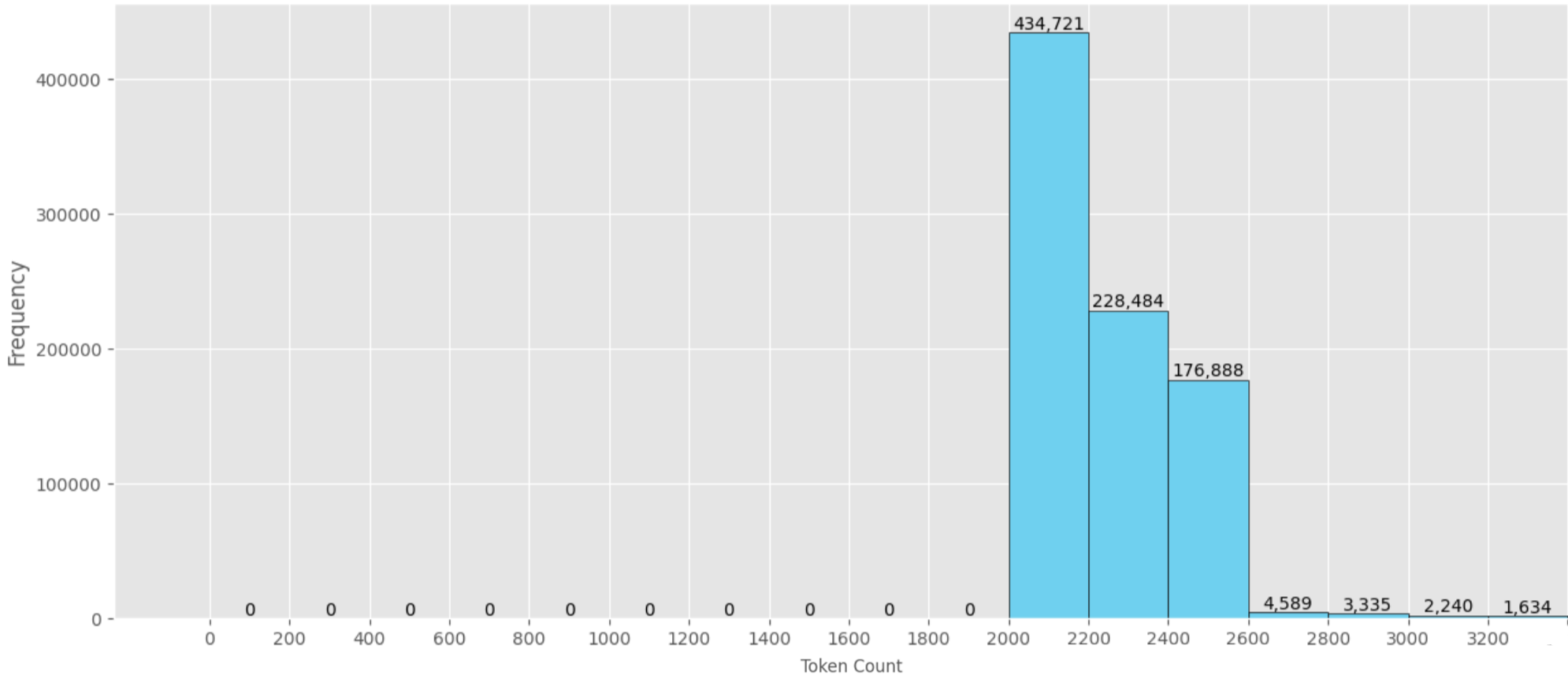}
        \caption{After packing}
        \label{fig:after_packing}
    \end{subfigure}

    \caption{Comparison of token count distributions per sample before and after packing.}
\end{figure}

Training on the OneVision dataset at scale poses efficiency challenges due to the bimodal length distribution of multi-image/video data and single-image data (see Figure~\ref{fig:before_packing}). A padding-based approach struggles with this heterogeneity, undermining training efficiency.

To address this issue, we employ an offline packing strategy, merging multiple instances (on average, three per pack) into a single input sequence, as inspired by~\cite{lu2023Unified,patchnpack}. To avoid interference across different cases within the same pack, we adjust the attention mask to ensure that attention is computed only within each individual case. By rebalancing the batch length distribution (see Figure~\ref{fig:after_packing}), packing achieves a 220\% improvement in training efficiency and substantially accelerates overall progress.

\vspace{-0.3cm}

\subsection{Annealing}
It is widely recognized that quality surpasses quantity in large multimodal models. Drawing inspiration from the annealing techniques introduced in MiniCPM~\cite{hu2024minicpm}, we apply curriculum learning and annealing during the instruction fine-tuning stage (Stage-2). We segment the instruction data from Stage-2 based on quality and defer higher-quality data to the latter part of training, where lower learning rates (Cosine LR Scheduler) are used. This annealing method yields stable and consistent performance gains.

\section{Experiment}
\label{exp}

\subsection{Evaluation on Open-source Benchmarks}
To rigorously evaluate our multi-modal model, Valley, we utilized eight benchmarks used in OpenCompass, encompassing diverse tasks and domains and providing a robust and comprehensive assessment of our model’s performance. These benchmarks include \textbf{MMBench}~\cite{MMBench}, which assesses the fine-grained capabilities of multi-modal models, ranging from perception to reasoning; \textbf{MMStar}~\cite{mmstar}, containing high-quality, manually reviewed samples that test the model’s general multi-modal capabilities; \textbf{MMMU}~\cite{yue2023mmmu}, designed to evaluate models on complex academic tasks; \textbf{MathVista}~\cite{lu2024mathvista}, focused on mathematical reasoning and problem-solving abilities; \textbf{HallusionBench}~\cite{HallusionBench}, measuring a model’s resistance to hallucinations and ensuring factual accuracy; \textbf{AI2D}~\cite{kembhavi2016diagramai2d}, which evaluates models’ understanding and reasoning abilities regarding natural science diagrams; \textbf{OCRBench}~\cite{ocrbench}, testing OCR and text comprehension in multi-modal contexts; and \textbf{MMVet}~\cite{yu2024mmvet}, assessing the model’s performance in comprehensive understanding and reasoning tasks.

The performance of Valley2 and other MLLMs on these benchmarks is shown in Table~\ref{tab:opencompass}. Valley2 ranks \textbf{No.2} in average score (\textbf{67.40}) among all models with fewer than 10B parameters, demonstrating its competitiveness across multiple tasks. Notably, our model achieves the highest score among all 10B models on the \textbf{MMMU} Benchmark, indicating that our model not only performs on par with or surpasses its peers in overall performance, but also excels in specific task domains, especially in complex academic tasks.

\begin{table}[ht]\small
\renewcommand\arraystretch{1.3}
\centering
\resizebox{1.0\textwidth}{!}{
\begin{tabular}{lccccccccc}
\toprule
\textbf{Models}         & \textbf{AVG} & \textbf{MB} & \textbf{MS} & \textbf{MMMU} & \textbf{Math} & \textbf{HB} & \textbf{AI2D} & \textbf{OCR} & \textbf{MV} \\ 
\midrule
SenseNova               & 77.40             & 85.7             & 72.7            & 69.6          & 78.4               & 57.4                    & 87.8          & 894              & 78.2           \\ 
Gemini-1.5-Pro-002      & 72.14            & 82.8             & 67.1            & 68.6          & 67.8               & 55.9                    & 83.3          & 770              & 74.6           \\ 
GPT-4o-20241120         & 72.02            & 84.3             & 65.1            & 70.7          & 59.9               & 56.2                    & 84.9          & 806              & 74.5           \\ 
\midrule
InternVL2.5-MPO-8B~\cite{wang2024mpo}          & 70.30            & 82             & \textbf{65.2}            & 54.8          & \textbf{67.9}              & \textbf{51.7}                      & 84.5       & \textbf{882}              & \textbf{68.1}           \\ 
POINTS1.5-7B~\cite{liu2024points1}    & 67.35            & 80.7             & 61.1            & 53.8          & 66.4               & 50                      & 81.4          & 832              & 62.2           \\ 
	
Qwen2-VL-7B~\cite{wang2024qwen2vl}    & 67          & 81           & 60.7         & 53.7          & 61.4              & 50.4                   & 83          & 843              & 61.8          \\ 
BlueLM-V-3B~\cite{2023bluelm}            & 66.11            & \textbf{82.7}             & 62.3            & 45.1          & 60.8               & 48                      & \textbf{85.3}          & 829              & 61.8           \\ 
MiniCPM-V-2.6~\cite{yao2024minicpm}          & 65.16            & 78               & 57.5            & 49.8          & 60.6               & 48.1                    & 82.1          & 852              & 60             \\ 
Ovis1.5-Llama3-8B~\cite{lu2024ovis}       & 62.25            & 76.6             & 57.3            & 48.3          & 63.0                 & 45                      & 82.5          & 744              & 50.9           \\ 
LLaVA-OV-7B (SI)~\cite{li2024llavaov} & 61.18            & 76.8             & 56.7            & 46.8          & 58.5               & 47.5                    & 82.8          & 697              & 50.6           \\ 
\midrule
Valley-7B~           & 67.40             & 80.68            & 60.93           & \textbf{57.0}           & 64.6               & 48.03                   & 82.48         & 842              & 61.28          \\ 
\bottomrule
\end{tabular}
}
\vspace{1mm}
\caption{Performance on Open-source Benchmarks. MB: MMbench, MS: MMStar, Math: MathVista, HB: HallusionBench, OCR: OCRBench, MV: MMVet.}
\label{tab:opencompass}
\end{table}

\subsection{Evaluation on Ecom-VQA Benchmark}

To comprehensively evaluate Valley’s performance in e-commerce scenarios, we conducted experiments on the Ecom-VQA Benchmark, comparing Valley2 with multimodal models of various sizes and scales. The results, shown in Table \ref{tab:ecom_vqa_results}., indicate that Valley2 outperforms other models with same scales in most tasks, achieving the highest overall average score. Notably, our 7B model even surpasses the best open-source 72B models in Single-Image test, including OCR and Image Understanding, and has also been validated across various real-world applications.

\begin{table}[ht!]\small
    \centering
    
    \renewcommand\arraystretch{1.3}
    \begin{tabular}{cm{0.7cm}m{0.7cm}m{0.7cm}m{0.7cm}m{0.7cm}m{0.7cm}m{0.7cm}m{0.7cm}m{0.7cm}m{0.7cm}}
        \toprule
        \multirow{2}{*}{Model} & \multirow{2}{*}{AVG} & \multicolumn{2}{c}{Image 120} & \multicolumn{3}{c}{Multi-image 179} & \multicolumn{4}{c}{Video 237} \\
        \cmidrule(lr){3-4} \cmidrule(lr){5-7} \cmidrule(lr){8-11}
        & & \makecell[c]{IU} & \makecell[c]{OCR} & \makecell[c]{DS} & \makecell[c]{MIU} & \makecell[c]{PTU}& \makecell[c]{GVU} & \makecell[c]{PEU} & \makecell[c]{AL} & \makecell[c]{VCR} \\
        \midrule
        InternVL2-72B & 77.80 &	90.00 &	78.33 &	90.00 &	93.22 &	81.67 &	93.33 &	91.53 &	50.00 	& 31.67    \\
        Qwen2-VL-72B & 86.06 & 93.33 & 83.33 & 88.33 & 93.22 & 83.33 & 96.67 & 91.53 & 55.17  & 88.33\\
        \midrule
        MiniCPM-V-2.6 & 55.56 & 73.33 & 43.30 & 58.33 & 77.97 & 50.00 & 78.33 & 67.80 & 1.72 & 3.33 \\
        InternVL2-8B & 63.25 & 88.33 & 66.66 & 70.00 & 86.44 & 71.67 & 91.67 & 77.97 & 10.34 & 5.00 \\
        LLaVA-OV-7B (SI)  & 66.04 & 85.00 & 75.00 & 70.00 & 89.83 & 78.33 & 83.33 & 81.36 & 15.52 & 15.00 \\
        Qwen2-VL-7B & 72.76 & 90.00 & 71.67 & 70.00 & 86.44 & 83.33 & 91.67 & 77.97 & 31.03 & 51.67 \\
        \midrule
        \textbf{Valley-7B}  & 79.66 & 95.00 & 86.67 & 80.00 & 89.83 & 81.67 & 93.33 & 88.14 & 50.00 & 51.67 \\
        \bottomrule
    \end{tabular}
    \vspace{1mm}
    \caption{Comparison with SoTA models on Ecom-VQA benchmark}
\label{tab:ecom_vqa_results}
\end{table}

\section{Ablation Study}
\subsection{Structure}

The Valley2 model incorporates four key structural components: Qwen2.5, ConvAdapter, Large MLP and the Eagle Module. To evaluate the contribution of each component, we conducted a series of ablation studies, with results summarized in Table~\ref{tab:ablation_structure}.  In the table, Ecom-caption evaluates the performance of the model after stage-1 training, which assesses the preliminary alignment of multimodal data. Ecom-VQA and OpenCompass evaluate the performance of the model after stage-2 training.

\textbf{Qwen2 vs.\ Qwen2.5.} Exps 2–3 highlight the substantial improvement of Qwen2.5 over Qwen2. This enhancement is especially pronounced in e-commerce-specific benchmarks and the OpenCompass benchmark. For instance, the Ecom-VQA score shows a notable increase from 74.58 to 82.35, indicating that Qwen2.5 possesses richer knowledge and superior reasoning capability, thus verifying its effectiveness in both language understanding and multimodal alignment tasks.

\textbf{PixelShuffle vs.\ ConvAdapter.} Exps 1–2 compare PixelShuffle and ConvAdapter. ConvAdapter introduces a lightweight parameterized structure to compress vision tokens along the spatial dimension. Specifically, assuming the visual feature dimension is $D_{\text{vis}}$, PixelShuffle (with a ratio of 2) concatenates four vision tokens along the channel dimension, resulting in an MLP input dimension of $4 \times D_{\text{vis}}$. In contrast, ConvAdapter merges these four tokens into one via a $2 \times 2$ convolution, preserving the dimension $D_{\text{vis}}$ and thereby reducing the MLP’s computational cost. The results show that ConvAdapter yields significant alignment benefits at Stage 1, boosting performance on the Ecom-Caption benchmark. This improvement propagates to the final results, delivering substantial gains on both the Ecom-VQA and OpenCompass benchmarks.

\textbf{Large MLP.} Exps 3–5 investigate the effect of enlarging the hidden size in the 2-layer MLP. The findings reveal that increasing the hidden size consistently improves performance on the Ecom-Caption benchmark, with a corresponding positive impact on end-to-end tasks, such as Ecom-VQA and OpenCompass. Ultimately, inspired by the Ovis design, we set the intermediate MLP dimension ($D_h$) to \nicefrac{1}{5} of the LLM vocabulary size and introduced a softmax activation function, achieving the best results on all three benchmarks. We also observed that tuning the learning rate is critical in this context. Empirical evidence suggests that if the MLP dimension is scaled by a factor of $n$, the learning rate should be scaled by $\sqrt{n}$. Future work will explore how activation functions and embedding distribution properties influence multimodal alignment, aiming to provide stronger interpretability.

\textbf{Eagle Module.} Exps 5–6 validate the effectiveness of the Eagle Module. Incorporating the vision encoder from Qwen2VL yielded a significant boost in OCR performance, as well as an approximate 2-point improvement on OpenCompass. However, integrating this module nearly doubles both training and inference time. During training, we limit the maximum image pixel count to constrain the number of vision tokens output by the Qwen2VL encoder. Nonetheless, at inference time, the module is highly scalable, allowing for further OCR performance gains by using higher-resolution images.

\begin{table}[ht]\small
    \centering
    \renewcommand\arraystretch{1.1} 
    \resizebox{1.0\textwidth}{!}{
    \begin{tabular}{c|cccccc|ccc}
        \toprule
        \multirow{2}{*}{Exp} & \multicolumn{2}{c}{LLM} & \multicolumn{2}{c}{\makecell[c]{Token Compress}} & \multirow{2}{*}{\makecell[c]{EAGLE\\Module}} & \multirow{2}{*}{\makecell[c]{MLP\\Hidden Size}} &  \multirow{2}{*}{E-Cap} &  \multirow{2}{*}{E-VQA} &  \multirow{2}{*}{OC} \\
        \cmidrule(lr){2-3} \cmidrule(lr){4-5} 
         & Qwen2 & Qwen2.5 & PixelShuffle & Conv &  &  &   &  & \\
        \midrule
        1 & \checkmark &  & \checkmark &  & & 1× & 35.01 & 64.55 & 51.03 \\
        2 & \checkmark &  &  & \checkmark & & 1× & 35.99 & 66.23 & 53.78 \\
        3 &  & \checkmark &  & \checkmark & & 1× & 36.35 & 73.13 & 59.37 \\
        4 &  & \checkmark &  & \checkmark & & 4× & 36.55 & 72.39 & 61.66 \\
        5 &  & \checkmark &  & \checkmark & & 9× & 36.97 & 74.44 & 63.15 \\
        6 &  & \checkmark &  & \checkmark & \checkmark & 9× & \textbf{37.45} & \textbf{79.66} & \textbf{66.1} \\
        \bottomrule
    \end{tabular}
    }
    \vspace{1mm}
    \caption{\textbf{Ablation of each component.} “Conv” refers to ConvAdapter; “MLP Hidden Size” by default is 3584; E-Cap: Ecom-Caption, E-VQA: Ecom-VQA, OC: OpenCompass.}
    \label{tab:ablation_structure}
\end{table}

\subsection{Annealing}
During the Instruction Fine-Tuning stage (Stage~2), we substantially enhanced the model’s instruction-following capability by employing a curriculum learning strategy rather than simple data mixing. Specifically, we divided the data into two subsets, where the second subset was markedly higher in sample quality, text length, and task complexity than the first. To verify the effectiveness of this curriculum-based approach, we compared two training strategies: (1) uniform mixed training and (2) annealed mixed training. As shown in Table~\ref{tab:annealing}, the annealed mixed strategy yielded substantial performance improvements across all metrics.

\subsection{Packing}
We performed offline packing by merging multiple samples into a single sample, ensuring that the final sequence lengths fell within a narrow range. This approach improved both training stability and speed but introduced variability in batch size. Nevertheless, as shown in Table~\ref{tab:packing}, offline packing did not cause any noticeable performance degradation at intermediate or final stages. Meanwhile, it significantly accelerated training, reducing total training time by a factor of approximately 2–3.

\subsection{Chain of Thought}
In Stage~3, we further incorporated the LLaVA-CoT-100k dataset\cite{xu2024llavacot} to enhance the model’s chain-of-thought (CoT) reasoning abilities. As described above, we appended the CoT prompt “Please think step by step” to the LLaVA-CoT-100k dataset and mixed it with 100,000 samples from the Stage~3 instruction data. This configuration allows the model to enter CoT mode simply by using the CoT prompt. The experimental results, presented in Table~\ref{tab:cot}, show that introducing CoT data improves the model’s performance on the OpenCompass benchmark.

\renewcommand{\arraystretch}{1.3}
\setlength{\tabcolsep}{2pt}
\begin{table}[ht]
    \renewcommand\arraystretch{1.3} 
    \centering
    \scalebox{0.91}{
        \begin{minipage}[t]{0.36\textwidth}
            \vspace{0pt} 
            \centering
            \begin{tabular}{p{1.4cm}<{\centering}p{1.4cm}<{\centering}p{1.4cm}<{\centering}}
                \toprule
                Anneal & E-VQA & OC \\
                \midrule
                 & 75.77 & 65.34 \\
                \checkmark & 79.66 & 66.11 \\
                \bottomrule
            \end{tabular}
            \vspace{3pt}
            \caption{Ablation on Annealing}
            \label{tab:annealing}
        \end{minipage}
        \hspace{0.1cm}
        
        \begin{minipage}[t]{0.31\textwidth}
            \vspace{0pt}
            \centering
            \begin{tabular}{p{1.2cm}<{\centering}p{1.2cm}<{\centering}p{1.0cm}<{\centering}p{1.2cm}<{\centering}}
                \toprule
                Packing & E-VQA & OC & Speed \\
                \midrule
                  & 79.16 & 66.20 & ×1 \\
                \checkmark &  79.66  & 66.11 & ×2.2 \\
                \bottomrule
            \end{tabular}
            \vspace{3pt}
            \caption{Ablation on Packing}
            \label{tab:packing}
        \end{minipage}
        \hspace{1cm}
        
        \begin{minipage}[t]{0.36\textwidth}
            \vspace{0pt}
            \centering
            \begin{tabular}{p{1.4cm}<{\centering}p{1.4cm}<{\centering}p{1.4cm}<{\centering}}
                \toprule
                CoT & E-VQA & OC \\
                \midrule
                 & 79.66  & 66.11 \\
                \checkmark & \textbf{} & \textbf{67.40} \\
                \bottomrule
            \end{tabular}
            \vspace{3pt}
            \caption{CoT Post-Training}
            \label{tab:cot}
        \end{minipage}
    }
    \label{tab:overall}
    \vspace{-4mm}
\end{table}
\section{Conclusion}
\label{conclusion}
We present the Valley2 series of multimodal models, consisting of both the Eagle and standard versions, with two open-source weight configurations of 8.9B and 8B parameters, respectively. Valley2 demonstrates outstanding performance across various multimodal scenarios while offering significant advantages in training and inference speed, as well as compatibility with diverse modalities. By exploring aggressive compression strategies for visual tokens, Valley2 can process longer video inputs and handle complex image-text-video tasks in mixed-modality settings. Thanks to its ConvAdapter-based projection structure and large visual embeddings, Valley2 consistently achieves competitive recognition and understanding performance, setting state-of-the-art results on both OpenCompass and in-house downstream tasks. Additionally, the Eagle variant provides scalable visual token support for extreme input scenarios, while the integration of Chain-of-Thought (CoT) reasoning further enhances performance in tasks requiring mixed capabilities.

Our overarching goal is to enable multimodal large models to be adopted in a wide range of industrial applications and at various budget levels, thus addressing numerous technical challenges. Guided by this vision, we continue to explore the frontiers of model architecture, data, and training methodologies.

\section{Coming Soon}
\label{Coming Soon}
To meet real-world application demands involving audio processing, retrieval tasks, and complex multimedia problem-solving, we are extending the Valley2 multimodal large language model in several directions. The following updates are planned for release in the near future:

\begin{enumerate}[label={\bf {{$\bullet$}}},,leftmargin=*,topsep=0.5ex,itemsep=-0.5ex,partopsep=0.75ex,parsep=0.75ex,partopsep=0pt,wide,labelindent=0pt]
    \item Developing an omni-model that integrates text, image, video, and audio modalities, with preliminary experiments achieving state-of-the-art results on audio-video understanding tasks.
    \item Introducing Valley-based multimodal embedding training approaches for downstream retrieval and probing applications across various modalities.
    \item Constructing complex benchmarks that rely on systematic CoT reasoning to evaluate both the individual and combined capabilities of multimodal models, combined with reinforcement learning to guide the development of more intelligent multimodal systems.
\end{enumerate}

\bibliographystyle{plainnat} 
\bibliography{references}

\clearpage
\appendix

\section{Detail of Ecom-VQA benchmark}
\begin{figure}[htbp]
    \centering
    \includegraphics[width=0.9\textwidth]{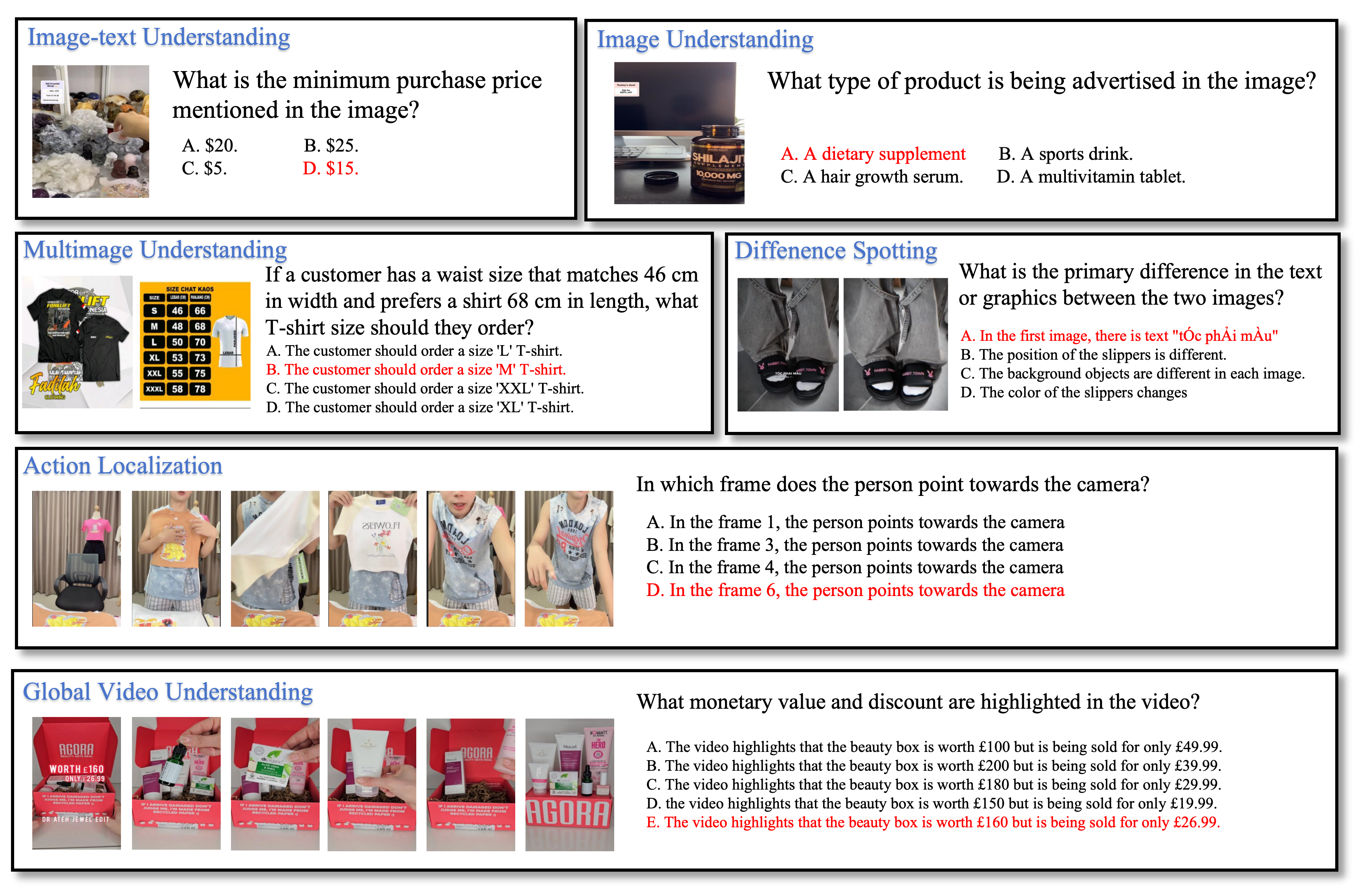} 
    \caption{Data of Ecom-VQA Benchmark}
    \label{fig:vqa_bar}
\end{figure}

\begin{table}[ht!]
\centering
\setlength{\tabcolsep}{12pt}
\renewcommand\arraystretch{1.2}
\begin{tabular}{p{3.5cm}p{9cm}}
\toprule
\textbf{Task} &  \textbf{Description} \\                                                                              
                                                                         
\specialrule{\heavyrulewidth}{0pt}{0pt} 
\multicolumn{2}{c}{Single Image}\\
\midrule

{\footnotesize Image Understanding}             & Understand the overall theme and main content of a single image and answer questions about the global semantics of the image.                                                                               \\
{\footnotesize Image-text understanding }                   & Answer related questions by recognizing text information within the image.\\
\midrule
\multicolumn{2}{c}{Multi Image}\\
\midrule
{\footnotesize Difference Spotting}  & Identify significant differences between two images, involving detail comparison and change detection, aimed at evaluating the ability to accurately detect and describe differences.\\
{\footnotesize Multi-image Understanding} & Accurately locate the time frame in a video where a specific action occurs, and answer detailed questions related to the spatial and temporal position of the action. \\
{\footnotesize Product Understanding}  & Understand the chronological order and causal relationships of events in a video, answering questions related to procedural steps and evaluating reasoning ability on procedural timelines.\\
\midrule
\multicolumn{2}{c}{Video}\\
\midrule
{\footnotesize Global Video Understanding}  & Understand the overall content of a video, analyze key themes, event to answer questions related to global information. \\
{\footnotesize Procedure Understanding } & Understand the chronological order and causal relationships of events in a video, answering questions related to procedural steps. \\
{\footnotesize Action Localization} & Accurately locate the time frame in a video where a specific action occurs. \\
{\footnotesize Video Content Retrieval}   & Determine whether the content of the given image appears in the given video.\\
\bottomrule

\end{tabular}
    \vspace{1mm}
\caption{Tasks and Descriptions for Each Modality.}
\label{tab:Ecom-VQA}
\end{table}

\end{document}